\documentclass{article}
\usepackage[preprint]{neurips_2026}

\usepackage[utf8]{inputenc}
\usepackage[T1]{fontenc}
\usepackage{hyperref}
\usepackage{url}
\usepackage{booktabs}
\usepackage{tabularx}
\usepackage{amsfonts}
\usepackage{amsmath}
\usepackage{nicefrac}
\usepackage{microtype}
\usepackage{xcolor}
\usepackage{graphicx}
\usepackage{subcaption}
\usepackage{float}
\usepackage{wrapfig}

\title{Privileged Context as Drift in On-Policy Self-Distillation}

\author{%
  Ravenor Davion\thanks{Equal contribution. Authors listed in random order.}\\
  Stanford University\\
  \texttt{rdavion@stanford.edu}
  \And
  Nick Rui\footnotemark[1]\\
  Stanford University\\
  \texttt{nickrui@stanford.edu}
}

\begin{document}

\maketitle

\begin{abstract}
On-policy self-distillation (OPSD) trains a language model to match a copy of itself conditioned on privileged context. Existing work varies what privileged context contains and how it is produced while also changing models, data, and training setups, making the effects of privileged context design difficult to isolate. Motivated by efforts in continual learning to reduce catastrophic forgetting, we study how the choice of privileged context affects policy drift. Specifically, we vary two axes: content (a demonstration, feedback, or rephrase) and source (external, self-generated with a verifier, or self-generated without a verifier). We train Qwen2.5-7B with OPSD across these nine combinations and three datasets, measuring target-task accuracy, prior-task retention, reverse KL from the base policy, and parameter-update geometry. Holding source fixed, changing content spans a wider median KL range than holding content fixed and changing source. The ratio between these ranges is $5.1\times$ for per-token KL and $2.2\times$ for per-sequence KL. Parameter-update geometry shows the same pattern: updates from adapters that share content are more closely aligned (mean cosine $0.571$) than updates from adapters that share source ($0.255$). For continual learning, these findings suggest that privileged context should be treated as part of OPSD's stability design because it is associated with how far and in what direction the policy moves.
\end{abstract}

\section{Introduction}
\label{sec:intro}
\begin{wrapfigure}[16]{r}{0.45\textwidth}
\vspace{-1.8\baselineskip}
\centering
\includegraphics[width=0.86\linewidth]{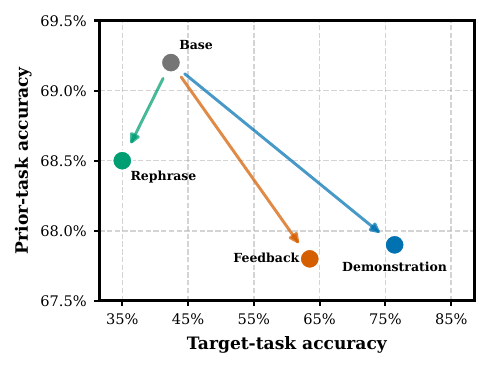}
\caption{Privileged context content choices are associated with different target-task and prior-task
outcomes. Points average over the three source conditions on ProofWriter; source-specific results
appear in Figure~\ref{fig:accuracy-retention}.}
\label{fig:proofwriter-performance-forgetting}
\vspace{-0.5\baselineskip}
\end{wrapfigure}
Recent frontier post-training pipelines increasingly separate \emph{capability acquisition} from \emph{capability integration}. Rather than jointly optimizing a single model across all target domains, these pipelines first train specialized policies using reinforcement learning and then consolidate their improvements into a unified model using on-policy distillation (OPD) \citep{deepseekai2026deepseekv4,xiaomi2026mimov2flash,nvidia2026nemotron3ultra,ma2026mopd}. This strategy addresses a central difficulty in large-scale post-training: optimizing a model for one domain can degrade capabilities acquired in others. Consequently, methods that can transfer improvements from specialized policies while limiting forgetting are becoming an increasingly important component of frontier model training.

These methods enable continual learning, where a model acquires new capabilities while preserving existing ones. \citet{shenfeld2025razor} show that policy drift, measured by KL divergence from the base policy, is strongly predictive of degradation in previously acquired capabilities, a phenomenon commonly referred to as catastrophic forgetting. Controlling unnecessary drift during post-training is therefore an essential objective for continual learning.

Recent work has proposed on-policy self-distillation (OPSD) as an approach to mitigating catastrophic forgetting \citep{shenfeld2026continual}. OPSD is designed to limit policy drift for two reasons. First, the teacher remains at the student's initialization and is conditioned on additional \emph{privileged context} \citep{vapnik2009privileged,lopezpaz2016unifying}. The initialized teacher anchors training targets to the base policy, while privileged context supplies the initial teacher--student disagreement. Second, OPSD is on-policy: training prefixes are sampled from the student, so teacher targets are evaluated on states that the student visits. This reduces the train--deployment state-distribution mismatch that can arise when training exclusively on fixed or teacher-generated sequences \citep{ross2011dagger,lin2020imitkd,agarwal2024gkd}.

This stability motivation makes privileged context a central design variable, as it determines the teacher's target distribution throughout training. We therefore study privileged context as drift, asking:

\begin{quote}\itshape
How do the content and source of privileged context affect policy drift in OPSD?
\end{quote}

Existing OPSD methods use a wide range of content as privileged context, including worked solutions \citep{op2sd,skald,ptdpo,unisd}, consensus answers \citep{canon,uopsd,coda}, verifier-accepted peer rollouts \citep{antisd,hsd}, execution traces \citep{vpd,hero}, self-critiques \citep{resd}, problem restatements \citep{msd,maigo,ccopd}, retrieved documents \citep{gates,latentmemcd}, rubrics \citep{rgsd,rcsd}, standing instructions \citep{crisp,opsa,bird}, compressed privileged context \citep{opsdl}, and learned soft tokens \citep{lopd}. We focus on three representative forms: \emph{demonstrations}, which provide an explicit solution to the query; \emph{feedback}, which provides diagnostics on a previous attempt; and \emph{rephrases}, which restate the query without providing an answer.

Privileged context also differs in its source. Some methods use privileged context generated by a larger or external model \citep{op2sd,skald,msd}, while others use privileged context generated by the student. Among self-generated privileged context artifacts, we distinguish between methods that use a correctness signal and those that do not. For example, some methods use a verifier to filter peer rollouts or trigger additional privileged context generation \citep{antisd,vpd,hsd}, whereas others generate privileged context without access to correctness feedback \citep{resd,maigo}. This distinction matters because access to correctness signals changes both the information available during training and the settings in which a method can be deployed. We therefore study source independently of content, considering three sources: an external source (a larger model), the student with verifier-based filtering, and the student without a verifier.

To disentangle these effects, we evaluate a full $3\times3$ content-by-source grid on three domains of tasks. We measure drift using per-token KL divergence from the base policy, cosine similarity between parameter updates, and changes in accuracy on prior-task benchmarks.

\section{Approach}
\label{sec:method}
\label{sec:design}

We isolate the effect of privileged context by crossing the three content types with the three sources while
holding base model, OPSD objective, optimization budget, and evaluation procedure fixed. The
result is nine conditions per dataset and 27 total trained adapters.

\subsection{Privileged Context Matrix}
\label{sec:approach:matrix}

Table~\ref{tab:context-matrix} summarizes the repository implementation of every cell. External
privileged context artifacts are generated by Qwen3.6-27B \citep{qwen2026qwen36}; both self-generated columns use a
frozen Qwen2.5-7B-Instruct \citep{qwen2024}. Gold answers are never included in a model prompt.
They are used only by the deterministic verifier in the External and Self $+$ Verifier columns.
The Self (no verifier) source uses a majority-vote-style plurality over sampled answers, with model
agreement in place of labels.

\begin{table}[t]
\centering
\small
\renewcommand{\arraystretch}{1.15}
\setlength{\tabcolsep}{4pt}
\begin{tabularx}{\textwidth}{@{}p{0.145\textwidth}
>{\raggedright\arraybackslash}X
>{\raggedright\arraybackslash}X
>{\raggedright\arraybackslash}X@{}}
\toprule
Content & External & Self $+$ Verifier & Self (no verifier) \\
\midrule
\textbf{Demonstration}
& Generate one Qwen3.6 worked solution; retain the query only when its extracted answer verifies.
& Sample eight frozen-Qwen2.5 solutions; among verifier-correct candidates, select the one with the highest mean token log-probability.
& Sample eight solutions, cluster their extracted answers, and select the highest-confidence solution in the plurality cluster. \\
\addlinespace[0.35em]
\textbf{Feedback}
& Critique one fixed Qwen2.5 attempt with Qwen3.6, apply the critique with frozen Qwen2.5, and retain it only when the correction verifies.
& Sample eight self-critiques of the fixed attempt, apply each, and select the critique producing the highest-confidence verifier-correct correction.
& Build a label-free consensus solution, sample eight critiques conditioned on it, apply each, and select by plurality agreement among the resulting corrections. \\
\addlinespace[0.35em]
\textbf{Rephrase}
& Generate one Qwen3.6 answer-preserving restatement and retain it only when frozen Qwen2.5 answers it correctly.
& Sample eight self-restatements, answer each with frozen Qwen2.5, and select the restatement producing the highest-confidence verifier-correct answer.
& Sample and answer eight self-restatements, cluster the downstream answers, and select the highest-confidence restatement in the plurality cluster. \\
\bottomrule
\end{tabularx}
\vspace{1em}
\caption{The $3\times3$ privileged context matrix. All artifacts are generated and selected
offline before training. During OPSD, the student always receives the original query; only the
frozen teacher receives the selected demonstration, attempt and feedback, or rephrased query.}
\label{tab:context-matrix}
\end{table}

Empty, truncated, unparseable, or verifier-rejected artifacts are rejected for that condition, and
the query is replaced from its registered reserve. Appendix~\ref{app:contexts} gives all nine privileged context artifacts for one
query, and Appendix~\ref{app:setup} gives the prompt and sampling details.

\subsection{Datasets}
\label{sec:approach:data}

We use ProofWriter OWA D5 at proof depths 4--5 \citep{tafjord2021proofwriter}, answerable MuSiQue
with three- and four-hop primary questions and a two-hop reserve used only for replacements
\citep{trivedi2022musique}, and hard Big-Math subsets \citep{albalak2025bigmath}. Big-Math uses a native held-out test
set matched to the training set's joint source and solve-rate profile (realized mean solve rate
$0.267$ versus $0.270$ in training).

\begin{table}[t]
\centering
\small
\begin{tabular}{lrrr}
\toprule
Dataset & Train & Dev. & Test \\
\midrule
ProofWriter OWA D5, depths 4--5 & 3{,}200 & 200 & 1{,}000 \\
MuSiQue answerable & 3{,}200 & 200 & 1{,}000 \\
Big-Math, hard subsets & 3{,}200 & 200 & 1{,}000 \\
\bottomrule
\end{tabular}
\vspace{1em}
\caption{Frozen dataset split sizes.}
\label{tab:splits}
\end{table}

Within each dataset, we audit candidates in one seed-0 primary-then-reserve order under all nine cells
and form a 3{,}200-query master set by descending number of cells in which privileged context construction
succeeds. Each condition keeps its accepted master queries and then, in the same order, tops up from
accepted candidates outside the master until it reaches 3{,}200. Table~\ref{tab:splits} reports the
split sizes. Task verifiers are exact-label match for ProofWriter, symbolic
equivalence through \texttt{math-verify} \citep{kydlicek2025mathverify} for Big-Math, and normalized
exact match against answer aliases for MuSiQue.

\subsection{Training}
\label{sec:approach:training}

We apply OPSD to Qwen2.5-7B-Instruct for each dataset and configuration, updating only a
student LoRA adapter while keeping the teacher frozen. Let $x_i$ be a query, $c_i$ its fixed privileged
context, and
$y_i\sim\pi_{\theta_s}(\cdot\mid x_i)$ a fresh rollout from the current pre-update student.
The frozen teacher and trainable student score the same rollout prefix,
$q_t^i(\cdot)=\pi_{\theta_0}(\cdot\mid x_i,c_i,y_i^{<t})$ and
$p_{\theta,t}^i(\cdot)=\pi_\theta(\cdot\mid x_i,y_i^{<t})$. For a batch $\mathcal B$, we
minimize the token-normalized full-vocabulary forward KL
\begin{equation}
\widehat{\mathcal L}(\theta)
=
\frac{1}{\sum_{i\in\mathcal B}T_i}
\sum_{i\in\mathcal B}\sum_{t=1}^{T_i}
\mathrm{KL}\!\left(q_t^i\,\middle\|\,p_{\theta,t}^i\right)
=
\frac{1}{\sum_{i\in\mathcal B}T_i}
\sum_{i\in\mathcal B}\sum_{t=1}^{T_i}\sum_{v\in\mathcal V}
q_t^i(v)\log\frac{q_t^i(v)}{p_{\theta,t}^i(v)} .
\label{eq:objective}
\end{equation}
Only the student LoRA parameters are updated; the same-weight teacher remains at
$\theta_0$ and is detached. Every run uses Qwen2.5-7B-Instruct with LoRA
\citep{hu2022lora} of rank 16, $\alpha=32$, and zero dropout on the
\texttt{q}, \texttt{k}, \texttt{v}, \texttt{o}, \texttt{gate}, \texttt{up}, and
\texttt{down} projections. We train for 861 constant-rate AdamW steps (learning rate
$10^{-5}$, weight decay $0.01$, effective batch size 16, microbatch size 1, gradient clipping 1.0,
BF16, and non-reentrant gradient checkpointing), yielding 13{,}776 fresh rollout exposures over
3{,}200 unique queries. The final step-861 checkpoint is fixed before test evaluation.

\subsection{Evaluation}
\label{sec:approach:evaluation}
\label{sec:design:drift}

We evaluate target-task accuracy, prior-task retention, distributional drift, and LoRA weight
geometry.

\paragraph{Task retention.}
Each adapter and the base model are evaluated with the Language Model Evaluation Harness
\citep{gao2024evalharness} on MMLU, HellaSwag, TruthfulQA-mc2, IFEval, and HumanEval
\citep{hendrycks2021mmlu,zellers2019hellaswag,lin2022truthfulqa,zhou2023ifeval,chen2021humaneval}.
For benchmark $b$, we compute
$\Delta_b=\mathrm{Acc}_{\mathrm{trained},b}-\mathrm{Acc}_{\mathrm{base},b}$ on the same items,
then report the unweighted mean of the five $\Delta_b$ values. Negative change is forgetting.
Per-item outcomes are paired when constructing intervals.

\paragraph{KL divergence.}
For every held-out prompt, we greedily decode both the trained and base models and teacher-force
both policies on each saved response. The primary drift measure is the full-vocabulary reverse KL
along the trained trajectory,
\begin{equation}
D_{\mathrm{rev}}
=
\sum_t
\mathrm{KL}\!\left(
\pi_{\mathrm{tr}}(\cdot\mid x,y_{\mathrm{tr}}^{<t})
\,\middle\|\,
\pi_{\mathrm{b}}(\cdot\mid x,y_{\mathrm{tr}}^{<t})
\right),
\label{eq:drift}
\end{equation}
averaged across prompts in nats per sequence and, separately, after within-prompt normalization in
nats per token. This pairing of reverse KL with trained-model prefixes follows the sequence-level
decomposition in prior work \citep{wen2023fdistill,zhao2026decoupling}. As a secondary response-length
diagnostic, Appendix~\ref{app:drift} reports the two-sample Kolmogorov--Smirnov statistic between
trained and base response lengths \citep{rabanser2019failing}.

\paragraph{Weight geometry.}
For every adapted module, the LoRA update is $\Delta W=(\alpha/r)BA$. We report the global
Frobenius norm
$\lVert\Delta W\rVert_F=(\sum_m\lVert\Delta W_m\rVert_F^2)^{1/2}$, the mean effective rank
$\exp(H(p))$ with $p_k=\sigma_k^2/\sum_j\sigma_j^2$, and the Frobenius cosine between the
vectorized updates of each condition pair. These measure update magnitude, spectral concentration,
and direction, respectively \citep{ilharco2023task,nagarajan2019distance,roy2007erank}.

\paragraph{Accuracy.}
Target-task accuracy uses greedy decoding on each frozen 1{,}000-example test set and the
dataset-specific verifier above. For every condition, we report base accuracy, trained accuracy,
their paired difference, and a 95\% paired-bootstrap interval with 10{,}000 resamples. These
intervals quantify test-item uncertainty; the experiment uses one training seed per condition.

\section{Results}
\label{sec:results}

\subsection{Base-Relative Comparisons}

Across all adapters, content choices span a wider range of distributional drift from the base policy
than source choices. Target-task accuracy and prior-task retention do not follow the same ordering.

\begin{figure}[H]
\centering
\includegraphics[width=\linewidth]{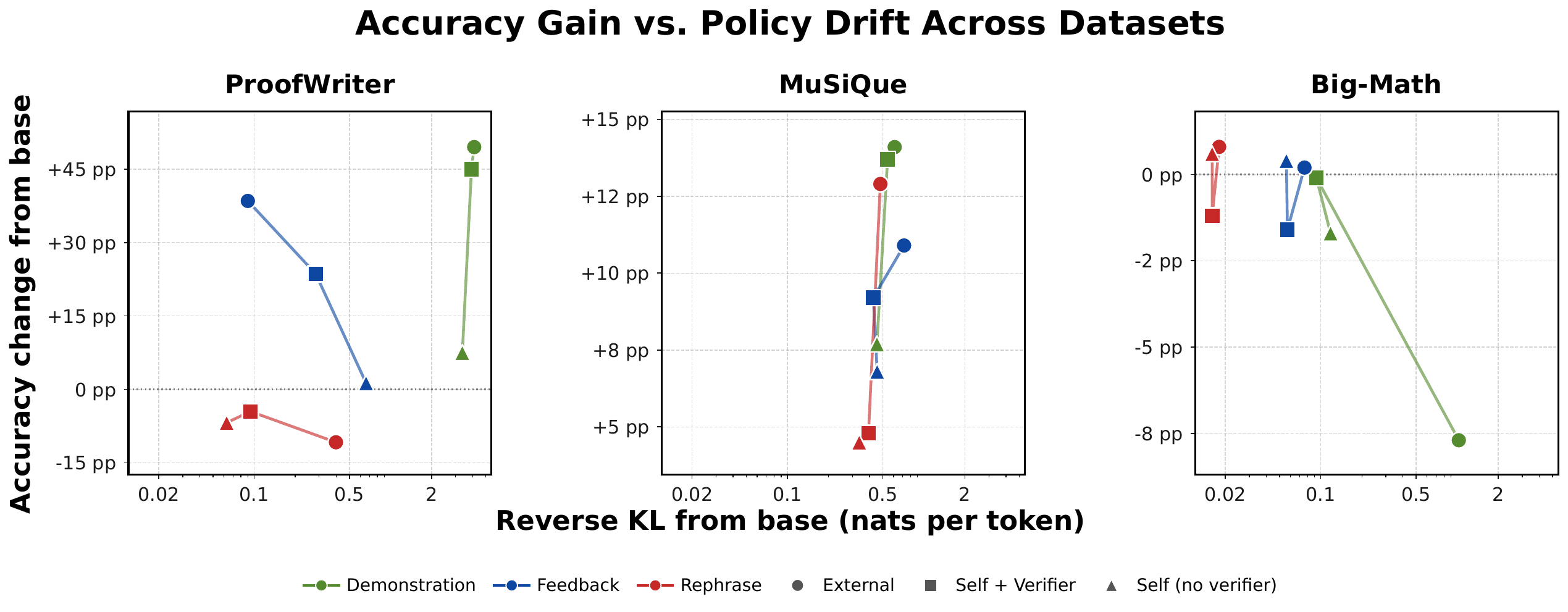}
\vspace{0.5em}
\caption{Target-task accuracy change and per-token distributional drift by privileged context
condition. Held-out accuracy change is plotted against trained-to-base reverse KL per token for one
adapter per condition. Each adapter uses one training seed. Color denotes content, marker denotes
source, and conditions with the same content are joined. The horizontal axis is logarithmic and
shared. Vertical scales differ.
Accuracy uses greedy decoding on 1{,}000 examples per dataset. Per-condition values and paired
bootstrap intervals are in Appendix Table~\ref{tab:percondition}.}
\label{fig:tradeoff}
\end{figure}

Figure~\ref{fig:tradeoff} plots target-task accuracy change against $D_{\mathrm{rev}}$ per token.
Holding source fixed, the median largest-to-smallest KL ratio across content choices is $7.2$ over
the nine dataset--source groups. Holding content fixed, the corresponding ratio across source
choices is $1.4$ over the nine dataset--content groups. Under per-sequence KL, the corresponding
median ratios are $2.80$ across content and $1.25$ across source. Thus, changing content spans a
wider median KL range under both summaries, although the ratio between the content and source
comparisons is $5.1$ per token and $2.2$ per sequence. These comparisons describe the selected
levels of each axis rather than general effect sizes, and one fixed-content group has an $11.1$-fold
per-token range across sources. Appendix~\ref{app:drift} reports the underlying per-condition
measurements.

Per-token KL ranges from $0.02$ nats for Big-Math rephrases to $4.10$ nats for ProofWriter
demonstrations. Demonstration and feedback conditions have positive target-task
accuracy point estimates for every source on ProofWriter and MuSiQue, with a maximum gain of $49.5$
points over the ProofWriter base accuracy of $42.4\%$. Rephrase conditions have positive point
estimates on MuSiQue and negative point estimates on ProofWriter. On Big-Math, eight of nine
target-task accuracy intervals include zero, and the External-demonstration condition decreases
accuracy by $7.7$ points. Larger KL from the base is therefore not consistently associated with
higher target-task accuracy.

Within most dataset--content groups, source choices span a narrower KL range, although target-task
accuracy can still differ. The ordering External $\geq$ Self $+$ Verifier $\geq$ Self is fully
monotonic for per-token KL in five of the nine dataset--content groups, has one pairwise inversion or
tie in three, and fully reverses in one. Target-task accuracy does not follow this ordering.
ProofWriter demonstrations, for example, span $42.0$ accuracy points across sources despite lying
within a $1.22$-fold per-token KL range.

Per-token and per-sequence KL also rank conditions differently. Their Spearman correlation is
$-0.148$ across adapters, while mean response length ranges from 5 to 622 tokens. ProofWriter
demonstrations have the largest per-token KL in the study, $4.10$ nats, but one of the smallest
per-sequence values, $20.6$ nats over a 5-token response. A Big-Math demonstration condition has
$0.09$ nats per token but accumulates $40.4$ nats over a 553-token response. Per-token KL measures
mean local divergence along the trained response, whereas per-sequence KL measures accumulated
divergence over the response. Because response length varies substantially across conditions, we
report both summaries and do not treat either as invariant to response length.

\begin{figure}[t]
\centering
\includegraphics[width=\linewidth]{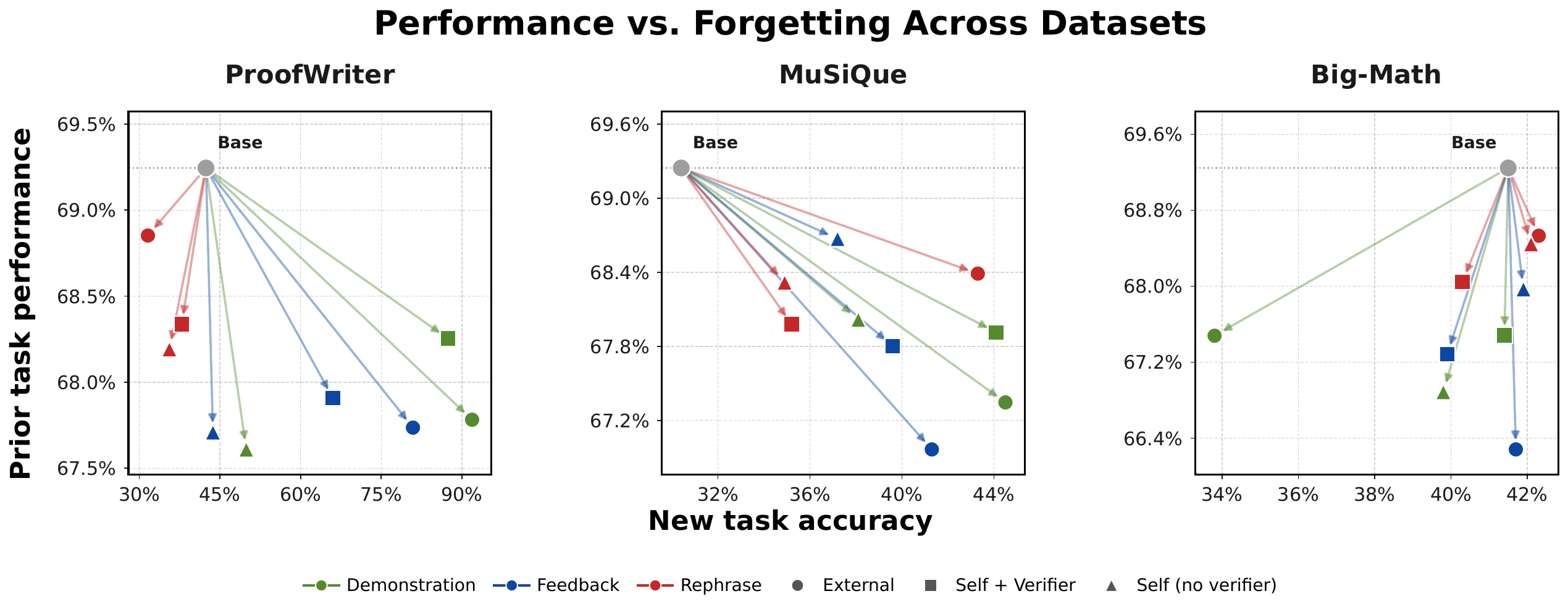}
\vspace{0.5em}
\caption{Target-task and prior-task accuracy by dataset and privileged context condition. Each arrow
starts at the base Qwen2.5-7B model (gray circle) and ends at one adapter. Each adapter uses one
training seed. The horizontal axis shows absolute held-out target-task accuracy. The vertical axis
shows the unweighted mean of absolute accuracy over five prior-task benchmarks (25{,}606 items).
Color denotes content and marker denotes
source. Both axes are scaled separately in each panel to make within-dataset differences visible.
Paired evaluation-item bootstrap intervals for prior-task accuracy changes are reported in Appendix
Table~\ref{tab:forgetting}; benchmark-level changes are in Appendix
Table~\ref{tab:forgetting-full}.}
\label{fig:accuracy-retention}
\end{figure}

Across all conditions, target-task accuracy changes by as much as $49.5$ points, whereas mean
prior-task accuracy changes by at most $3.0$ points (Figure~\ref{fig:accuracy-retention}). Most
conditions show a decrease in prior-task accuracy with a paired evaluation-item bootstrap interval
that excludes zero. The largest benchmark-level decreases occur on IFEval and HumanEval, at
$8.3$ and $6.1$ points, respectively, while MMLU changes by no more than $0.21$ points across
conditions. Because each condition uses one training seed, these intervals quantify only
evaluation-item uncertainty. Appendix~\ref{app:forgetting} reports the condition- and benchmark-level
changes.

Figure~\ref{fig:proofwriter-performance-forgetting} shows source-averaged ProofWriter outcomes as a
compact illustration. These averages mask substantial variation within content: demonstration
accuracy spans $42.0$ points across sources. We therefore do not interpret the separation among
source-averaged conditions as an isolated content effect; the source-specific results in
Figure~\ref{fig:accuracy-retention} form the basis of the held-fixed comparisons.

Task retention does not follow the content--source pattern observed for per-token KL. Most paired
source contrasts in prior-task accuracy have intervals that include zero. Among the remaining
contrasts, four oppose the KL-derived ordering and one follows it. Across adapters, per-token KL has
Spearman correlation $-0.188$ with prior-task accuracy change, and target-task accuracy change has
correlation $-0.053$ with prior-task accuracy change. The largest observed association is between
per-sequence KL and prior-task accuracy change, with correlation $-0.480$.

\subsection{Pairwise Adapter Comparisons}

\begin{figure}[t]
\centering
\begin{subfigure}[b]{0.285\linewidth}
\includegraphics[width=\linewidth]{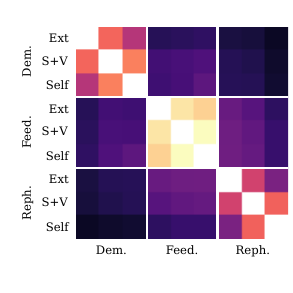}
\caption{ProofWriter}
\end{subfigure}
\hfill
\begin{subfigure}[b]{0.285\linewidth}
\includegraphics[width=\linewidth]{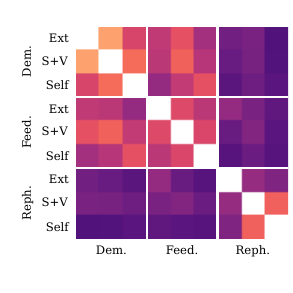}
\caption{MuSiQue}
\end{subfigure}
\hfill
\begin{subfigure}[b]{0.285\linewidth}
\includegraphics[width=\linewidth]{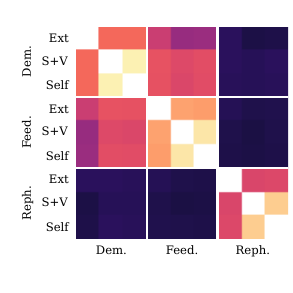}
\caption{Big-Math}
\end{subfigure}

\vspace{0.4em}
\includegraphics[width=0.42\linewidth]{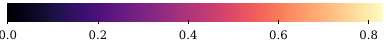}
\vspace{1em}
\caption{Pairwise LoRA-update cosine similarity by dataset and privileged context condition. Each
panel shows Frobenius cosine between the LoRA deltas of the nine single-seed adapters trained on one
dataset, ordered by content and then source.}
\label{fig:cosine}
\end{figure}

Holding source fixed and changing content gives a mean update cosine of $0.255$
(Figure~\ref{fig:cosine}). Holding content fixed and changing source gives a mean cosine of $0.571$.
For reference, updates trained on different datasets have mean cosine $0.034$ over 243 pairs. Within
a dataset, adapters trained with the same content and different sources therefore have more closely
aligned update directions than adapters trained with the same source and different content.

The two self-generated sources are the most closely aligned source pair in seven of the nine
dataset--content groups. Their mean cosine is $0.662$, compared with $0.563$ for External versus
Self $+$ Verifier and $0.488$ for External versus Self. Self $+$ Verifier is closer to External than
Self is in eight groups and closer to Self than External is in seven. These comparisons often place
Self $+$ Verifier between the other source conditions, but they do not establish that the three
sources lie on a single trajectory.

\section{Discussion}
\label{sec:discussion}

\subsection{Analysis}

The base-relative and pairwise comparisons measure different properties, yet both distinguish
content from source in the same way. With source fixed, changing content gives median
largest-to-smallest KL ratios of $7.2$ per token and $2.80$ per sequence, along with a mean
parameter-update cosine of $0.255$. With content fixed, changing source gives corresponding median
KL ratios of $1.4$ and $1.25$ and a mean cosine of $0.571$. Content choices are therefore associated
with larger differences in drift magnitude under both KL summaries and with larger differences in
update direction within this grid, although the size of the KL contrast depends on aggregation.

Adapter geometry remains associated with source after distillation. External and Self $+$ Verifier
both use verifier-based selection, yet Self $+$ Verifier is closer to Self in seven of nine
dataset--content groups and closer to External than Self is in eight. This pattern is compatible
with generator-specific properties persisting after distillation and with prior evidence that
distillation transfers teacher mechanisms incompletely \citep{wu2024mechanisms}, although that work
studies teacher mechanisms rather than generators of privileged context. However, the External
condition also changes model identity, capacity, prompting, temperature, and selection procedure,
so the experiment cannot isolate that explanation or place the three sources on a single trajectory.

Target-task accuracy and task retention do not follow the same content--source contrast. ProofWriter
demonstrations span $42.0$ target-task accuracy points across sources within a $1.22$-fold per-token KL range, while
most Big-Math conditions change the output distribution without a detectable accuracy gain. The
weak rank correlations among per-token KL, target-task accuracy change, and prior-task accuracy
change likewise do not support treating any one measure as a proxy for the others. The three
measurements describe acquisition, distributional drift, and retention separately.

Several limits bound this interpretation. The content and source levels differ in kind, so the
held-fixed KL comparisons depend on the selected levels, and their magnitude depends on whether
divergence is aggregated per token or per sequence. Because response length varies with content
type, neither summary is length-invariant, and part of the content-associated KL range may reflect
response length rather than per-step distributional change. All 27 adapters share one base model, LoRA rank,
training budget, and seed; evaluation-item bootstrap intervals do not measure training variance, so smaller contrasts may
not replicate across seeds, and repeating the grid over multiple seeds is needed to confirm them.
ProofWriter and MuSiQue use overlapping but non-identical training-query sets across conditions,
whereas Big-Math uses a shared set. The KL estimates follow greedy trajectories. Finally, the
External source confounds authorship with model identity and capacity, so source differences
cannot be attributed to generator capacity or verification individually, and the grid excludes other
forms of privileged context such as retrieved documents, rubrics, and standing instructions.

\subsection{Implications for Continual Learning}

For continual post-training, privileged context is part of the stability design. The same-model
teacher anchors the training target to the initialized policy, and on-policy rollouts obtain teacher
targets on prefixes sampled from the student. The use of model outputs as stability targets parallels
Learning without Forgetting, which distills the previous model on new-task inputs while fitting the new task
\citep{li2016learning}. OPSD differs by conditioning its teacher on privileged context and evaluating
the teacher on student rollouts. Neither property fixes the teacher's conditional distribution,
which is determined by privileged context. The resulting adapters can therefore drift by different
amounts and in different directions even when the model, objective, optimization budget, and
on-policy sampling procedure are held fixed.

This distinction matters when OPSD is used to integrate specialized capabilities. A condition can
produce a large target-task gain with limited prior-task change, or alter the output distribution
without improving the target task. Selecting privileged context only by target-task accuracy can
therefore obscure differences in retention and distributional drift. Evaluating acquisition,
retention, and drift separately gives a more complete account of whether an OPSD update adds a
capability while limiting movement from the base policy.

\section{Conclusion}
\label{sec:conclusion}

We study how the content and source of privileged context affect OPSD outcomes while holding the model,
training objective, optimization budget, and evaluation procedure fixed. Across three datasets,
changing content is associated with larger differences in drift from the base policy and in
parameter-update direction than changing source. Source differences are smaller but not absent: adapters
trained under the two self-generated conditions generally have more closely aligned updates with
each other than with the External condition. These patterns agree across base-relative and pairwise
comparisons, but they do not identify a universally preferable content or source. Target-task
acquisition, prior-task retention, and distributional drift often rank the same conditions
differently.

Taken together, these findings motivate treating privileged context as part of the stability design
of OPSD. Same-model distillation and on-policy sampling constrain the training setup, but they do
not determine how far or in what direction the student moves. OPSD systems intended for continual
learning should therefore evaluate acquisition, retention, and drift separately rather than
select privileged context using target-task accuracy alone.

It remains open which properties of privileged context generators account for the source-associated
differences observed here. More granular, controlled comparisons that separate authorship, model
capacity, prompting, and verifier-based selection could help identify their respective contributions.

\bibliographystyle{plainnat}
\bibliography{references}

\appendix
\section{Implementation Details}
\label{app:setup}

\subsection{Generation}

Self-generation draws eight candidates at temperature $1.0$ and top-$p$ $0.95$. The external
model runs at temperature $0.7$ with thinking mode disabled. Ties in any selection rule are
broken by mean token log-probability. Any empty, unparseable, truncated, or verifier-rejected
artifact is treated as rejected and the query is replaced from a prespecified reserve, so no
condition trains on a failed artifact.

\subsection{Optimization}

We use AdamW at a constant learning rate of $10^{-5}$, with weight decay $0.01$, gradient
clipping $1.0$, effective batch size 16 with microbatch size 1, BF16, and non-reentrant
gradient checkpointing.
LoRA is applied to all attention and MLP projections at rank 16, $\alpha = 32$, dropout 0.
Each run draws a fresh on-policy rollout every time a query is selected, giving 13{,}776 rollout
exposures over 861 optimizer steps. All runs use seed $0$. Diagnostic checkpoints are written at
steps 215, 431, 646, and 861; step 861 was fixed in advance as the reported checkpoint and no
checkpoint is selected using test performance.

\subsection{Evaluation}

Task accuracy is greedy decoding on the frozen 1{,}000-example test split, with the prespecified
per-dataset verifier and identical response limits for base and trained models. Intervals are
paired bootstrap over test examples with $B = 10{,}000$ resamples. Prior-task benchmarks are run
at full size under the Language Model Evaluation Harness, at 14{,}042 items for MMLU, 10{,}042
for HellaSwag, 817 for TruthfulQA-mc2, 541 for IFEval, and 164 for HumanEval.

\section{Privileged Context Examples}
\label{app:contexts}

This is a quality-screened illustration, not a random sample. Among the 637 query identifiers
present in all nine final ProofWriter condition datasets, we selected one with three
verifier-correct demonstrations, a verifier-incorrect fixed attempt, and three self-contained
rephrases whose selected downstream labels match the task label. All generated wording is
reproduced verbatim, including Markdown markup. The feedback prompt requests a targeted critique rather than a replacement solution but
does not prohibit an explicit answer label. Selection checks the downstream correction rather than
each claim in the feedback; the Self feedback therefore retains its inconsistent final line.

\subsection{Query}
\begin{quote}\small\ttfamily
Erin is kind. Erin is nice. Erin is quiet. Erin is round. Fiona is cold. Fiona is kind. Fiona is nice. Fiona is quiet. Gary is green. Harry is green. Harry is quiet. Harry is round. If Erin is cold then Erin is quiet. If someone is young and round then they are quiet. If someone is young and cold then they are kind. If someone is young and green then they are kind. If someone is green then they are young. All kind, cold people are quiet. Quiet people are nice. If someone is quiet then they are young. All kind, green people are cold.

Based on the information above, is the following statement True, False, or Unknown? Gary is not quiet.
\end{quote}

\subsection{Demonstration, External}
\begin{quote}\small\ttfamily
To determine whether the statement "Gary is not quiet" is True, False, or Unknown, we must deduce the properties of Gary based on the provided facts and rules.

**1. Analyze the Facts regarding Gary:**
*   Fact: Gary is green.

**2. Analyze the Rules:**
*   Rule 1: If Erin is cold then Erin is quiet.
*   Rule 2: If someone is young and round then they are quiet.
*   Rule 3: If someone is young and cold then they are kind.
*   Rule 4: If someone is young and green then they are kind.
*   Rule 5: If someone is green then they are young.
*   Rule 6: All kind, cold people are quiet.
*   Rule 7: Quiet people are nice.
*   Rule 8: If someone is quiet then they are young.
*   Rule 9: All kind, green people are cold.

**3. Step-by-Step Deduction for Gary:**

*   **Step 1:** Determine if Gary is young.
    *   From Fact: Gary is green.
    *   From Rule 5: "If someone is green then they are young."
    *   Since Gary is green, **Gary is young**.

*   **Step 2:** Determine if Gary is kind.
    *   From Step 1: Gary is young.
    *   From Fact: Gary is green.
    *   From Rule 4: "If someone is young and green then they are kind."
    *   Since Gary is young and green, **Gary is kind**.

*   **Step 3:** Determine if Gary is cold.
    *   From Step 2: Gary is kind.
    *   From Fact: Gary is green.
    *   From Rule 9: "All kind, green people are cold."
    *   Since Gary is kind and green, **Gary is cold**.

*   **Step 4:** Determine if Gary is quiet.
    *   From Step 3: Gary is cold.
    *   From Step 2: Gary is kind.
    *   From Rule 6: "All kind, cold people are quiet."
    *   Since Gary is kind and cold, **Gary is quiet**.

    *Alternative path for Step 4:*
    *   From Step 1: Gary is young.
    *   From Fact: Gary is green.
    *   Does Rule 2 apply? "If someone is young and round then they are quiet." We know Gary is young, but we do not know if Gary is round. So this rule doesn't directly apply yet.
    *   However, the deduction in Step 4 using Rule 6 is sufficient and definitive. Gary is quiet.

**4. Evaluate the Query:**
*   The query asks: "Gary is not quiet."
*   Our deduction shows: **Gary is quiet.**
*   Therefore, the statement "Gary is not quiet" contradicts the deduced fact.

**Conclusion:**
The statement is False.

Final answer: False
\end{quote}

\subsection{Demonstration, Self $+$ Verifier}
\begin{quote}\small\ttfamily
To determine whether the statement "Gary is not quiet" is True, False, or Unknown, let's analyze the given facts and rules step by step.

1. **Facts and Rules**:
   - Erin is kind.
   - Erin is nice.
   - Erin is quiet.
   - Erin is round.
   - Fiona is cold.
   - Fiona is kind.
   - Fiona is nice.
   - Fiona is quiet.
   - Gary is green.
   - Harry is green.
   - Harry is quiet.
   - Harry is round.
   - If Erin is cold then Erin is quiet. (Not relevant as Erin is not cold.)
   - If someone is young and round then they are quiet.
   - If someone is young and cold then they are kind.
   - If someone is young and green then they are kind.
   - If someone is green then they are young.
   - All kind, cold people are quiet.
   - Quiet people are nice.
   - If someone is quiet then they are young.
   - All kind, green people are cold.

2. **Analyzing Gary**:
   - Gary is green.
   - From the rule "If someone is green then they are young," we can infer that Gary is young.
   - From the rule "If someone is young and green then they are kind," since Gary is both young and green, Gary is kind.
   - From the rule "All kind, green people are cold," since Gary is kind and green, Gary is cold.
   - From the rule "If someone is young and cold then they are kind," we confirm that Gary is kind (already known).
   - From the rule "All kind, cold people are quiet," since Gary is both kind and cold, Gary is quiet.

Since we can logically conclude from the given facts and rules that Gary is quiet, the statement "Gary is not quiet" is False.

Final answer: False
\end{quote}

\subsection{Demonstration, Self}
\begin{quote}\small\ttfamily
To determine whether the statement "Gary is not quiet" is True, False, or Unknown, let's analyze the given facts and rules step by step.

1. **Facts and Rules**:
   - Erin is kind.
   - Erin is nice.
   - Erin is quiet.
   - Erin is round.
   - Fiona is cold.
   - Fiona is kind.
   - Fiona is nice.
   - Fiona is quiet.
   - Gary is green.
   - Harry is green.
   - Harry is quiet.
   - Harry is round.
   - If Erin is cold then Erin is quiet. (Not relevant as Erin is not cold.)
   - If someone is young and round then they are quiet.
   - If someone is young and cold then they are kind.
   - If someone is young and green then they are kind.
   - If someone is green then they are young.
   - All kind, cold people are quiet.
   - Quiet people are nice.
   - If someone is quiet then they are young.
   - All kind, green people are cold.

2. **Analyzing Gary**:
   - Gary is green.
   - From the rule "If someone is green then they are young," we can infer that Gary is young.
   - From the rule "If someone is young and green then they are kind," since Gary is both young and green, Gary is kind.
   - From the rule "All kind, green people are cold," since Gary is kind and green, Gary is cold.
   - From the rule "If someone is young and cold then they are kind," we confirm that Gary is kind (already known).
   - From the rule "All kind, cold people are quiet," since Gary is both kind and cold, Gary is quiet.

Since we can logically conclude from the given facts and rules that Gary is quiet, the statement "Gary is not quiet" is False.

Final answer: False
\end{quote}

\subsection{Fixed Attempt}
The same verifier-incorrect attempt is supplied to all three feedback generators.
\begin{quote}\small\ttfamily
To determine if the statement "Gary is not quiet" is true, false, or unknown, let's analyze the given information step by step.

1. **Gary's Characteristics:**
   - Gary is green.
   - Gary is quiet.

2. **Rules and Inferences:**
   - "If someone is green then they are young." (Gary is green, therefore Gary is young.)
   - "If someone is young and green then they are kind." (Gary is young and green, therefore Gary is kind.)
   - "If someone is young and green then they are kind." (Repetition, not needed.)
   - "All kind, green people are cold." (Since Gary is both young and green, and "All kind, green people are cold," this implies Gary is cold.)
   - "All kind, cold people are quiet." (Since Gary is both kind and cold, this implies Gary is quiet.)

3. **Check Contradictions:**
   - We do not have a rule or given information stating that Gary cannot be quiet, nor do we have a direct contradiction to Gary being quiet.

4. **Deductions:**
   - Since all the conditions that imply Gary should be quiet are satisfied (young and green, and thus kind and cold), we cannot say for certain that Gary is not quiet based solely on the given information.

5. **Conclusion:**
   - Given the information, Gary being quiet is a necessary outcome from the provided rules. Thus, we do not have a basis to assert that Gary is not quiet.

Final answer: Unknown
\end{quote}

\subsection{Feedback, External}
\begin{quote}\small\ttfamily
The error lies in the final classification of the answer. You correctly deduced that Gary **is** quiet based on the rules (Green \$\textbackslash{}rightarrow\$ Young \$\textbackslash{}rightarrow\$ Kind \$\textbackslash{}rightarrow\$ Cold \$\textbackslash{}rightarrow\$ Quiet). Since the statement "Gary is not quiet" directly contradicts the derived fact that Gary is quiet, the statement is **False**, not Unknown. Under the open-world assumption, a statement is False only if its negation is entailed by the premises; here, the premises entail that Gary is quiet, so the claim that he is not quiet is false.
\end{quote}

\subsection{Feedback, Self $+$ Verifier}
\begin{quote}\small\ttfamily
The first important error is in the conclusion. The statement "Since all the conditions that imply Gary should be quiet are satisfied, we cannot say for certain that Gary is not quiet based solely on the given information" is incorrect. The information states that if someone is young and green, they are kind and, therefore, cold, and all kind, cold people are quiet. Since Gary is young and green, you can deduce that Gary is quiet.

Concrete change needed: Since Gary is young and green, and all kind, cold people are quiet, you can conclude that Gary is quiet. Therefore, the statement "Gary is not quiet" is False.

Final answer: False
\end{quote}

\subsection{Feedback, Self}
\begin{quote}\small\ttfamily
The analysis correctly deduces that Gary is quiet based on the given rules. However, the conclusion should state that the statement "Gary is not quiet" is False, not Unknown.

Final answer: Unknown (should be False)
\end{quote}

\subsection{Rephrase, External}
\begin{quote}\small\ttfamily
Erin is kind, nice, quiet, and round. Fiona is cold, kind, nice, and quiet. Gary is green. Harry is green, quiet, and round. Consider the following rules:
1. If Erin is cold, then Erin is quiet.
2. If someone is young and round, then they are quiet.
3. If someone is young and cold, then they are kind.
4. If someone is young and green, then they are kind.
5. If someone is green, then they are young.
6. All people who are both kind and cold are quiet.
7. All quiet people are nice.
8. If someone is quiet, then they are young.
9. All people who are both kind and green are cold.

Based on this information, is the statement "Gary is not quiet" True, False, or Unknown?
\end{quote}

\subsection{Rephrase, Self $+$ Verifier}
\begin{quote}\small\ttfamily
Erin is kind. Erin is nice. Erin is quiet. Erin is round. Fiona is cold. Fiona is kind. Fiona is nice. Fiona is quiet. Gary is green. Harry is green. Harry is quiet. Harry is round. If Erin is cold then Erin is quiet. If someone is young and round then they are quiet. If someone is young and cold then they are kind. If someone is young and green then they are kind. If someone is green then they are young. All kind, cold people are quiet. Quiet people are nice. If someone is quiet then they are young. All kind, green people are cold.

Is the statement "Gary is not quiet" True, False, or Unknown?
\end{quote}

\subsection{Rephrase, Self}
\begin{quote}\small\ttfamily
Erin is kind. Erin is nice. Erin is quiet. Erin is round. Fiona is cold. Fiona is kind. Fiona is nice. Fiona is quiet. Gary is green. Harry is green. Harry is quiet. Harry is round. If Erin is cold then Erin is quiet. If someone is young and round then they are quiet. If someone is young and cold then they are kind. If someone is young and green then they are kind. If someone is green then they are young. All kind, cold people are quiet. Quiet people are nice. If someone is quiet then they are young. All kind, green people are cold.

Is the statement "Gary is not quiet" True, False, or Unknown?
\end{quote}

\clearpage
\section{Prior-Task Accuracy}
\label{app:forgetting}

\begin{table}[H]
\centering
\small
\begin{tabular}{lccc}
\toprule
Content & External & Self $+$ Verifier & Self \\
\midrule
\multicolumn{4}{l}{\textit{ProofWriter}} \\
\quad Demonstration & $-$0.0146$^{*}$ & $-$0.0099$^{*}$ & $-$0.0164$^{*}$ \\
\quad Feedback & $-$0.0151$^{*}$ & $-$0.0134$^{*}$ & $-$0.0154$^{*}$ \\
\quad Rephrase & $-$0.0039 & $-$0.0091$^{*}$ & $-$0.0105$^{*}$ \\
\multicolumn{4}{l}{\textit{MuSiQue}} \\
\quad Demonstration & $-$0.0190$^{*}$ & $-$0.0133$^{*}$ & $-$0.0123$^{*}$ \\
\quad Feedback & $-$0.0228$^{*}$ & $-$0.0144$^{*}$ & $-$0.0057 \\
\quad Rephrase & $-$0.0086 & $-$0.0127$^{*}$ & $-$0.0093$^{*}$ \\
\multicolumn{4}{l}{\textit{Big-Math}} \\
\quad Demonstration & $-$0.0177$^{*}$ & $-$0.0176$^{*}$ & $-$0.0236$^{*}$ \\
\quad Feedback & $-$0.0296$^{*}$ & $-$0.0196$^{*}$ & $-$0.0128$^{*}$ \\
\quad Rephrase & $-$0.0071 & $-$0.0120$^{*}$ & $-$0.0080 \\
\bottomrule
\end{tabular}
\vspace{1em}
\caption{Mean prior-task accuracy change by dataset and privileged context condition. Entries are changes
in the unweighted mean accuracy over five benchmarks (25{,}606 items).
$^{*}$ marks a paired 95\% bootstrap interval over evaluation items that excludes zero. Each
adapter uses one training seed, whose variance is not represented by the interval.}
\label{tab:forgetting}
\end{table}

Table~\ref{tab:forgetting} reports the suite mean, aggregating without weights over the five
benchmark-level changes reported below for every condition.

\begin{table}[H]
\centering
\footnotesize
\setlength{\tabcolsep}{4pt}
\begin{tabular}{llrrrrr}
\toprule
Content & Source & MMLU & HellaSwag & TruthfulQA & IFEval & HumanEval \\
\midrule
\multicolumn{7}{l}{\textit{ProofWriter}} \\
\quad Demonstration & External & $+0.01$ & $-0.04$ & $-0.61$$^{*}$ & $-2.40$ & $-4.27$$^{*}$ \\
\quad Demonstration & Self $+$ Verifier & $+0.14$ & $-0.19$$^{*}$ & $+0.12$ & $-2.59$ & $-2.44$ \\
\quad Demonstration & Self & $+0.00$ & $-0.14$ & $-0.13$ & $-4.25$$^{*}$ & $-3.66$ \\
\quad Feedback & External & $+0.05$ & $-0.24$$^{*}$ & $-0.59$$^{*}$ & $-5.55$$^{*}$ & $-1.22$ \\
\quad Feedback & Self $+$ Verifier & $+0.09$ & $-0.17$$^{*}$ & $-0.65$$^{*}$ & $-3.51$$^{*}$ & $-2.44$ \\
\quad Feedback & Self & $-0.07$ & $-0.19$$^{*}$ & $-0.36$ & $-4.62$$^{*}$ & $-2.44$ \\
\quad Rephrase & External & $+0.05$ & $-0.03$ & $-0.39$$^{*}$ & $-0.37$ & $-1.22$ \\
\quad Rephrase & Self $+$ Verifier & $-0.16$ & $-0.15$$^{*}$ & $-0.61$$^{*}$ & $-2.40$ & $-1.22$ \\
\quad Rephrase & Self & $-0.19$$^{*}$ & $-0.10$ & $-0.45$$^{*}$ & $-1.48$ & $-3.05$ \\
\midrule
\multicolumn{7}{l}{\textit{MuSiQue}} \\
\quad Demonstration & External & $-0.05$ & $-0.10$ & $-2.29$$^{*}$ & $-4.62$$^{*}$ & $-2.44$ \\
\quad Demonstration & Self $+$ Verifier & $-0.02$ & $+0.02$ & $-1.82$$^{*}$ & $-2.40$ & $-2.44$ \\
\quad Demonstration & Self & $-0.12$ & $-0.10$ & $-1.13$$^{*}$ & $-2.96$ & $-1.83$ \\
\quad Feedback & External & $-0.04$ & $-0.15$ & $-2.01$$^{*}$ & $-5.55$$^{*}$ & $-3.66$ \\
\quad Feedback & Self $+$ Verifier & $-0.09$ & $-0.12$ & $-0.69$ & $-3.88$$^{*}$ & $-2.44$ \\
\quad Feedback & Self & $-0.08$ & $-0.10$ & $-0.92$$^{*}$ & $-0.55$ & $-1.22$ \\
\quad Rephrase & External & $-0.21$ & $-0.20$$^{*}$ & $-1.73$$^{*}$ & $-0.92$ & $-1.22$ \\
\quad Rephrase & Self $+$ Verifier & $+0.01$ & $-0.10$ & $-0.79$$^{*}$ & $-2.40$ & $-3.05$ \\
\quad Rephrase & Self & $-0.01$ & $-0.09$ & $-0.75$$^{*}$ & $-0.74$ & $-3.05$ \\
\midrule
\multicolumn{7}{l}{\textit{Big-Math}} \\
\quad Demonstration & External & $+0.05$ & $-0.19$$^{*}$ & $-0.41$ & $-4.62$$^{*}$ & $-3.66$ \\
\quad Demonstration & Self $+$ Verifier & $-0.09$ & $-0.13$ & $-0.50$ & $-4.44$$^{*}$ & $-3.66$ \\
\quad Demonstration & Self & $-0.15$ & $-0.20$$^{*}$ & $-0.47$ & $-6.10$$^{*}$ & $-4.88$$^{*}$ \\
\quad Feedback & External & $-0.04$ & $-0.27$$^{*}$ & $-0.09$ & $-8.32$$^{*}$ & $-6.10$$^{*}$ \\
\quad Feedback & Self $+$ Verifier & $-0.18$ & $-0.17$$^{*}$ & $-0.01$ & $-5.18$$^{*}$ & $-4.27$$^{*}$ \\
\quad Feedback & Self & $-0.07$ & $-0.19$$^{*}$ & $-0.09$ & $-4.81$$^{*}$ & $-1.22$ \\
\quad Rephrase & External & $-0.09$ & $-0.09$ & $+0.03$ & $-0.37$ & $-3.05$ \\
\quad Rephrase & Self $+$ Verifier & $-0.21$ & $-0.14$ & $-0.34$ & $+0.18$ & $-5.49$$^{*}$ \\
\quad Rephrase & Self & $-0.19$ & $+0.01$ & $-0.24$ & $+1.29$ & $-4.88$$^{*}$ \\
\bottomrule
\end{tabular}
\vspace{1em}
\caption{Prior-task accuracy change by benchmark and privileged context condition, in percentage points.
$^{*}$ marks a
paired 95\% bootstrap interval over evaluation items that excludes zero; all adapters use one
training seed.}
\label{tab:forgetting-full}
\end{table}

\clearpage
\section{Per-Condition Measurements}
\label{app:drift}

We report every quantity used in the main text for all 27 conditions. Base accuracy is given per
dataset in the header rows. The two divergence columns report the divergence in
Eq.~\eqref{eq:drift} per token and per sequence. Length is the trained
model's mean response length in tokens, and erank is effective rank, the mean over adapted
modules out of 16.

\begin{table}[H]
\centering
\footnotesize
\setlength{\tabcolsep}{2.2pt}
\begin{tabular}{llrrrrrrrr}
\toprule
Content & Source & $\Delta$acc & 95\% CI & /tok & /seq & KS & len & $\lVert \Delta W \rVert_F$ & erank \\
\midrule
\multicolumn{10}{l}{\textit{ProofWriter}, base 0.424} \\
\quad Demonstration & External & $+0.495$ & $[+0.46,+0.53]$ & $4.104$ & $20.6$ & $1.00$ & $5$ & $1.75$ & $6.78$ \\
\quad Demonstration & Self $+$ Verifier & $+0.450$ & $[+0.41,+0.49]$ & $3.920$ & $19.6$ & $1.00$ & $5$ & $2.25$ & $6.12$ \\
\quad Demonstration & Self & $+0.075$ & $[+0.04,+0.11]$ & $3.361$ & $16.8$ & $1.00$ & $5$ & $2.38$ & $6.10$ \\
\quad Feedback & External & $+0.385$ & $[+0.35,+0.42]$ & $0.090$ & $25.7$ & $0.15$ & $336$ & $3.18$ & $5.40$ \\
\quad Feedback & Self $+$ Verifier & $+0.236$ & $[+0.20,+0.27]$ & $0.283$ & $23.1$ & $0.15$ & $281$ & $2.95$ & $5.82$ \\
\quad Feedback & Self & $+0.013$ & $[-0.02,+0.05]$ & $0.662$ & $22.7$ & $0.27$ & $228$ & $2.94$ & $5.94$ \\
\quad Rephrase & External & $-0.108$ & $[-0.14,-0.07]$ & $0.398$ & $15.9$ & $0.29$ & $234$ & $2.56$ & $6.25$ \\
\quad Rephrase & Self $+$ Verifier & $-0.045$ & $[-0.08,-0.01]$ & $0.094$ & $8.9$ & $0.16$ & $280$ & $2.15$ & $7.01$ \\
\quad Rephrase & Self & $-0.068$ & $[-0.10,-0.03]$ & $0.063$ & $8.1$ & $0.18$ & $283$ & $2.26$ & $6.96$ \\
\midrule
\multicolumn{10}{l}{\textit{MuSiQue}, base 0.304} \\
\quad Demonstration & External & $+0.141$ & $[+0.11,+0.17]$ & $0.611$ & $4.5$ & $0.43$ & $8$ & $2.44$ & $5.71$ \\
\quad Demonstration & Self $+$ Verifier & $+0.137$ & $[+0.11,+0.17]$ & $0.542$ & $4.1$ & $0.41$ & $8$ & $2.41$ & $5.88$ \\
\quad Demonstration & Self & $+0.077$ & $[+0.05,+0.10]$ & $0.453$ & $3.6$ & $0.40$ & $9$ & $2.53$ & $6.01$ \\
\quad Feedback & External & $+0.109$ & $[+0.08,+0.14]$ & $0.716$ & $6.5$ & $0.38$ & $10$ & $2.76$ & $5.56$ \\
\quad Feedback & Self $+$ Verifier & $+0.092$ & $[+0.06,+0.12]$ & $0.426$ & $3.7$ & $0.38$ & $10$ & $2.45$ & $6.19$ \\
\quad Feedback & Self & $+0.068$ & $[+0.04,+0.09]$ & $0.455$ & $3.7$ & $0.39$ & $9$ & $2.57$ & $6.23$ \\
\quad Rephrase & External & $+0.129$ & $[+0.10,+0.15]$ & $0.480$ & $13.7$ & $0.19$ & $45$ & $2.71$ & $5.43$ \\
\quad Rephrase & Self $+$ Verifier & $+0.048$ & $[+0.02,+0.07]$ & $0.393$ & $3.9$ & $0.33$ & $12$ & $2.37$ & $6.29$ \\
\quad Rephrase & Self & $+0.045$ & $[+0.02,+0.07]$ & $0.336$ & $4.6$ & $0.22$ & $22$ & $2.32$ & $6.51$ \\
\midrule
\multicolumn{10}{l}{\textit{Big-Math}, base 0.415} \\
\quad Demonstration & External & $-0.077$ & $[-0.10,-0.05]$ & $1.028$ & $40.1$ & $0.59$ & $240$ & $3.32$ & $6.00$ \\
\quad Demonstration & Self $+$ Verifier & $-0.001$ & $[-0.02,+0.02]$ & $0.093$ & $40.4$ & $0.17$ & $553$ & $2.91$ & $6.42$ \\
\quad Demonstration & Self & $-0.017$ & $[-0.04,+0.01]$ & $0.118$ & $43.5$ & $0.22$ & $523$ & $2.92$ & $6.36$ \\
\quad Feedback & External & $+0.002$ & $[-0.02,+0.03]$ & $0.076$ & $41.2$ & $0.09$ & $622$ & $3.08$ & $6.35$ \\
\quad Feedback & Self $+$ Verifier & $-0.016$ & $[-0.04,+0.01]$ & $0.057$ & $29.7$ & $0.12$ & $592$ & $2.98$ & $6.56$ \\
\quad Feedback & Self & $+0.004$ & $[-0.02,+0.03]$ & $0.056$ & $28.7$ & $0.15$ & $590$ & $2.99$ & $6.52$ \\
\quad Rephrase & External & $+0.008$ & $[-0.01,+0.03]$ & $0.018$ & $8.2$ & $0.09$ & $609$ & $2.47$ & $7.56$ \\
\quad Rephrase & Self $+$ Verifier & $-0.012$ & $[-0.04,+0.01]$ & $0.016$ & $7.1$ & $0.10$ & $601$ & $2.31$ & $7.83$ \\
\quad Rephrase & Self & $+0.006$ & $[-0.02,+0.03]$ & $0.016$ & $7.2$ & $0.10$ & $601$ & $2.28$ & $7.83$ \\
\bottomrule
\end{tabular}
\vspace{1em}
\caption{Task accuracy, distributional drift, response length, update norm, and effective rank by
privileged context condition. All 27 adapters use one training seed. Accuracy intervals are paired
bootstrap intervals over evaluation items and do not include training-seed variance.}
\label{tab:percondition}
\end{table}


\end{document}